\documentclass[twoside]{article}

\usepackage[accepted]{aistats2025}
 \usepackage{natbib} 
 \usepackage[colorlinks=true, citecolor=blue, linkcolor=blue, urlcolor=blue]{hyperref}
\usepackage{titletoc}
\usepackage{xcolor}  % 引入 xcolor 包

\usepackage{bbding}
\usepackage{arydshln}
\hypersetup{
    colorlinks=true,   % 开启颜色链接
    linkcolor=blue     % 设置 \ref 的颜色为蓝色
}
\usepackage[full]{textcomp}
\usepackage{url}
\usepackage{algorithm}  % 如果使用 \KwIn, \KwOut 命令
\usepackage{algpseudocode}
\usepackage{amsmath}
\usepackage{float}
\usepackage{amsmath, amssymb}
\usepackage{graphicx}
\usepackage{booktabs}
\usepackage{multirow}

\usepackage{natbib}
\usepackage{hyperref}
\usepackage{url}
\usepackage{lipsum}
\usepackage{graphicx} %
\usepackage{amsmath} %
\usepackage{placeins}
\usepackage{amsfonts} %
\usepackage{amsthm} %
\usepackage{amssymb} %
\usepackage{bbold} %
\usepackage[only,llbracket,rrbracket]{stmaryrd}
\usepackage{amsthm} %
\usepackage{booktabs}
\usepackage{multirow}
\usepackage{caption}
\usepackage{subcaption}
\usepackage{textcomp} 
\usepackage{stix}

\definecolor{gbcolor}{rgb}{0.9,0,0.5}

\makeatletter%罗马数字

\newcommand{\Rmnum}[1]{\expandafter\@slowromancap\romannumeral #1@}
\begin{document}



\twocolumn[

\aistatstitle{Stochastic Weight Sharing for Bayesian Neural Networks}

\aistatsauthor{ Moule Lin \And Shuhao Guan \And  Weipeng Jing \And Goetz Botterweck \And Andrea Patane }

\aistatsaddress{ Lero, Trinity College \\ Dublin \And  University College \\ Dublin \And Northeast Forestry \\ University \And Lero, Trinity College \\Dublin \And Lero, Trinity College \\ Dublin} ]

\begin{abstract}
While offering a principled framework for uncertainty quantification in deep learning, the employment of Bayesian Neural Networks (BNNs) is still constrained by their increased computational requirements and the convergence difficulties when training very deep, state-of-the-art architectures.
In this work, we reinterpret weight-sharing quantization techniques from a stochastic perspective in the context of training and inference with Bayesian Neural Networks (BNNs). Specifically, we leverage 2D-adaptive Gaussian distributions, Wasserstein distance estimations, and alpha-blending to encode the stochastic behaviour of a BNN in a lower-dimensional, soft Gaussian representation. Through extensive empirical investigation, we demonstrate that our approach significantly reduces the computational overhead inherent in Bayesian learning by several orders of magnitude, enabling the efficient Bayesian training of large-scale models, such as ResNet-101 and Vision Transformer (VIT). On various computer vision benchmarks—including CIFAR-10, CIFAR-100, and ImageNet1k—our approach compresses model parameters by approximately 50× and reduces model size by 75\%, while achieving accuracy and uncertainty estimations comparable to the state-of-the-art.

% while only sacrificing 2\% accuracy on the ImageNet1k dataset. Additionally, it facilitates approximate Bayesian training of large models such as ResNet-50, ResNet-100, and Vision Transformers,
\end{abstract}

\section{INTRODUCTION}

Bayesian Neural Networks (BNNs) promise to combine the representational capacity of deep learning with principled uncertainty estimations enabled by means of Bayesian learning theory \citep{neal2012bayesian}. Arguably, this combination makes them particularly appealing for safety-critical machine learning applications where the quantification of uncertainty is of paramount importance \citep{forsberg2020challenges}. Indeed, they have been widely employed in scenarios like e-Health \citep{marcos2010classification}, robust control \citep{wicker2024probabilistic}, autonomous driving \citep{michelmore2020uncertainty}, human-in-the-loop applications \citep{treiss2021uncertainty}, automated diagnosis \citep{billah2022bayesian} and many others \citep{lampinen2001bayesian,bharadiya2023review,vehtari1999bayesian}. 

Unfortunately, though, the principled treatment of uncertainty comes at the price of an increased pressure on computational resources, including the model size ($\times2$ in the common case of mean-field Variational Inference \citep{blundell2015weight}), and the inference time, increased by an order of magnitude as multiple forward passes are needed \citep{neal2012bayesian}.
Therefore, despite their potential, the use of BNNs in edge-AI and resource-constrained applications is still very limited \citep{bonnet2023bringing}. 
%Furthermore, even in more standard applications, BNNs are generally limited to shallower architectures than their deterministic counter-parts \citep{goan2020bayesian,izmailov2021bayesian, dusenberry2020efficient,su2020sampling}. 
While recent works have investigated the development of techniques to tackle the aforementioned challenges, these are generally limited to the application of methods originally developed for deterministic neural networks (NNs) \citep{ferianc2021effects,park2021vector,chien2023bayesian}, or the usage of the Bayesian paradigm at training time for model-order reduction purposes but without the uncertainty estimation at inference time \citep{van2020bayesian,guo2018survey,perrin2024hardware,subia2024probabilistic}.

In this work, we present a quantisation technique specifically tailored to capture the stochastic behaviour of BNNs.
More specifically, we design a stochastic weight-sharing quantisation method, called 2DGBNN, based on dynamically adaptive mini-batch 2D Gaussian Mixture Models and predicated on optimising weight distributions through metrics derived from Wasserstein distances \citep{chizat2020faster,de2021quantum}, network gradients, and intra-class variance. Our technique works by reinterpreting standard weight-sharing \citep{subia2024probabilistic} from a 2D perspective, accounting for both the mean and variance of BNN's parameters in the case of mean-field Variational Inference (VI), and by giving it a stochastic semantic. At each training step, the current sum-total of network parameters is clustered using standard parameter-free techniques, and a mini-batch approach is used for the estimation of Gaussian distributions on a parameter space accounting for (possibly) millions of parameters. Representatives for each cluster are then selected, and alpha-blending techniques are used for sampling parameter realisations during forward passes through the network architecture. Thanks to its simplicity, the method can be seamlessly integrated into commonly used Bayesian approximation techniques based on Variational Inference \citep{blundell2015weight,gal2016dropout,minka2001expectation,welling2011bayesian}. 

We perform an extensive empirical investigation on the effectiveness of our method in training large-scale models and in reducing the computational footprint of BNNs. We utilize four widely used image classification datasets (i.e., MNIST~\citep{lecun1998gradient}, CIFAR-10~\citep{krizhevsky2009learning}, CIFAR-100~\citep{krizhevsky2009learning} and ImageNet1K~\citep{deng2009imagenet}) and test the model results in 
four widely employed neural network architectures, including ResNet-18, ResNet-50, ResNet-101~\citep{he2016deep} and {Vision Transformer (ViT)}~\citep{dosovitskiy2021image}. Through our approach, we can reduce the number of trainable parameters in the network by up to $50\times$, while obtaining accuracy and uncertainty metrics comparable to the state-of-the-art.\footnote{To support reproducibility, our code is available at {\url{https://github.com/moulelin/2DGBNN}}}

This paper makes the following main contributions:
\begin{itemize}
    \item We introduce a stochastic weight-sharing technique specifically-tailored to BNNs and that employs Wasserstein distance, gradients, and within-class variance in order to improve model efficiency.
    \item We empirically demonstrate that our stochastic weight-sharing method compares favourably against quantisation methods employed for BNNs in terms of further reducing the computational requirements and better preserving accuracy and uncertainty metrics. 
    \item In a variety of architectures, including ResNet-18, ResNet-50, ResNet-10 and ViT, we show how our technique can reduce model size by up to just a quarter of the original size, while working on par with state-of-the-art techniques for large-scale approximate BNN training. 
\end{itemize}
\section{RELATED WORK}
%\begin{itemize}
    %\item quantisation for BNNs \citep{ferianc2021effects,park2021vector,chien2023bayesian,subedar2021quantization,lin2023quantization,yang2020variational}.  
    %\item Bayesian techniques for quantisation of NNs \citep{subia2024probabilistic,louizos2017bayesian,van2020bayesian,perrin2024hardware,achterhold2018variational,soudry2014expectation}
    %\item Hardware acceleration for BNNs [to have in intro as interesting references rather than in related work] \citep{fan2022fpga,cai2018vibnn,bonnet2023bringing}
    %\item Additional references of interest \citep{zhou2021efficient}
%\end{itemize}
Computational efficiency is one of the long-standing issues concerning the application of BNNs to edge-AI and embedded systems \citep{bonnet2023bringing}. Indeed, there is a vast section of the literature that aims to tackle the issue from a variety of different angles. 
One such closely related area is that of quantisation of the BNN's parameters \citep{guo2018survey}, where the precision of the latter is reduced to minimise their computational footprint. Works have adapted techniques initially developed for deterministic NNs~\citep{ferianc2021effects,subedar2021quantization,lin2023quantization,dong2022finding,ullrich2017soft} or developed new techniques that take into account the distributional behaviour of BNNs' parameters \citep{chien2023bayesian,park2021vector,yang2020variational}. 
%, and evaluated them in terms of accuracy, uncertainty estimation, and computational footprint.
While these techniques increase the computational efficiency of a given BNN architecture, by developing a weight-sharing technique  (and therefore not only quantising but also in effect reducing the number of BNN parameters) the method we introduce is able to match their behaviour in standard BNN benchmarks, while at the same time, it allows for training of large scale models  \citep{hernandez2015probabilistic}. 

Several works have looked at reducing the number of parameters in BNNs, either by applying pruning techniques \citep{sharma2021bayesian,beckers2023principled,roth2018bayesian} or using low-rank approximations \citep{doan2024bayesian,dusenberry2020efficient,swiatkowski2020k}. %Concerning the former set of techniques, while through pruning has proved that models size can be reduced of up to 80%80\% without significantly impacting the accuracy \citep{sharma2021bayesian}, parameters are simply cut off by pruning techniques. In contradistinction, our method allows one to reduce the model size while at the same time keeping the influence of all networks parameters through the use of sharing distributions. 
The latter work by reparameterising BNN weights and biases using a lower rank representation, enabling them to scale BNN inference to large models such as ResNet-50 \citep{dusenberry2020efficient} and ViT \citep{doan2024bayesian} architectures, and are as such closely related to our work in that a smaller common representation is found for BNN parameters. The number of final parameters is still though generally higher than their full deterministic counterpart, and therefore cannot be used for edge-AI applications. In Section \ref{sec:experiments}, we will observe that, albeit at the price of a small reduction in accuracy, our method reduces the number of parameters of 3 orders of magnitude.

%On weight sharing (but not to BNNs) \citep{roth2018bayesian,nowlan2018simplifying}...

Finally, a number of works have looked into applying Bayesian techniques for quantisation of deterministic NNs \citep{subia2024probabilistic,louizos2017bayesian,van2020bayesian,perrin2024hardware,achterhold2018variational,soudry2014expectation,yang2020searching}, including the application of weight-sharing techniques \citep{roth2018bayesian,subia2024probabilistic,nowlan2018simplifying}. While these techniques provide encouraging results on the suitability of Bayesian theory for increasing the efficiency of deep learning, being specifically tailored to deterministic neural networks they cannot be applied to BNNs.

%\section{Background}
%In this section, we briefly review the two key components that make up this paper's methodology, Bayesian Neural Networks (BNNs) and weight-sharing quantisation.

\section{BAYESIAN NEURAL NETWORKS}
%BNNs combine the representational power of neural networks with the principled approach of Bayesian inference \citep{neal2012bayesian, jospin2022hands}, providing a probabilistic framework \citep{hernandez2015probabilistic,charnock2022bayesian} that quantifies uncertainty in model predictions. 
We consider a neural network architecture $f^\mathbf{w} : \mathbb{R}^n \rightarrow \mathbb{R}^m$ parameterised by a vector of weights and biases $\mathbf{w} \in \mathbb{R}^{n_w}$, which we refer to collectively as parameters of the neural network. Bayesian learning of neural networks begins by placing a prior distribution, $p(\mathbf{w})$, over the networks' parameter. This is often assumed to be encoded through a vector of independent Gaussian distributions, one for each weight and bias in the BNN \citep{blundell2015weight}. This prior belief is then updated given a dataset's evidence through the application of the Bayesian learning rule. Let $\mathcal{D} = \{(x_i,y_i)\}_{i=1}^{n_{\mathcal{D}}}$ denote the full training dataset, $\mathbf{X} = (x_1,\ldots,x_{n_{\mathcal{D}}})$ the combined vector of training inputs and $\mathbf{y} = (y_1,\ldots,y_{n_{\mathcal{D}}})$ their corresponding outputs, then the posterior distribution on the weight is computed as:
\begin{align} \label{eq:posterior}
    p(\mathbf{w}|\mathbf{X}, \mathbf{y}) = \frac{p(\mathbf{y}|\mathbf{X}, \mathbf{w}) \, p(\mathbf{w})}{p(\mathbf{y}|\mathbf{X})},
\end{align}
where $p(\mathbf{y}|\mathbf{X}, \mathbf{w})$ is the likelihood and $p(\mathbf{y}|\mathbf{X})$ is the model evidence. Finally, given a test point $x^*$, the BNN's posterior predictive distribution on $x^*$ is defined by:
\begin{align} \label{eq:predictive}
p(y^* | x^* , \mathbf{X}, \mathbf{y}) = \int p(y^* | x^*, \mathbf{w}) p(\mathbf{w} | \mathbf{X}, \mathbf{y} )d\mathbf{w}.
\end{align}
Unfortunately, neither Equation \eqref{eq:posterior} nor Equation \eqref{eq:predictive} can generally be computed exactly \citep{neal2012bayesian}. Therefore a variety of approximate Bayesian inference techniques have been developed in the literature, with the two most prominent classes of approaches being based on Monte Carlo algorithms \citep{neal2012bayesian} or on Variational Inference methods \citep{blundell2015weight}. 
While the former provides the gold standard in terms of approximation accuracy in small architectures, Variational Inference (VI) guarantees better scalability and is therefore the focus of this paper.
Briefly, VI works by approximating the true posterior, \( p(\mathbf{w}|\mathbf{X}, \mathbf{y})\), by optimising the KL divergence over a simpler, parameterised distribution, \( q(\mathbf{w}) \), often a multidimensional Gaussian distribution with diagonal covariance. %\AP{If ELBO is needed we can add its details in here, but very briefly.} 
The predictive distribution of Equation \eqref{eq:predictive} is then approximated by sampling multiple times from \( q(\mathbf{w}) \), and averaging the results. 

Despite the approximation, however, VI BNNs still come with several limitations that impede their deployment in practice. First, even in the case of diagonal Gaussian distributions, the number of parameters in the BNN is doubled compared to their deterministic counterpart. Furthermore, approximating the predictive distribution requires multiple sampling procedures and multiple forward passes through the network so that their computational time is orders of magnitude higher than, again, their deterministic counterpart. Finally, despite its greater flexibility, standard Variational Inference struggles to learn very deep BNNs and it is generally limited to more traditional architecture and small-to-medium-size datasets. In the following, we develop a weight-sharing quantisation scheme targeted at BNNs to tackle these limitations.

\section{2D GAUSSIAN BAYESIAN NEURAL NETWORK}
\label{sec:methods}

Consider the vector $\mathbf{w}$ of the BNN's weights, where, at each step of the training process, each weight, $w_i$ $i=1,\ldots,n_{\mathbf{w}}$, is distributed accordingly to a given Gaussian distribution $\mathcal{N}(\mu_{w_i},\sigma_{w_i})$. We denote with $\mathcal{N}_{full}$ the full set of weight distributions. Our stochastic weight sharing technique aims at finding a set of 2D Gaussian distributions (which we collectively denote as $\mathcal{N}_{\textrm{ws}}$) $\mathcal{N}(\mu_1,\Sigma_1), \ldots, \mathcal{N}(\mu_k,\Sigma_k)$,\footnote{As standard we use $\sigma$ to denote the one-dimensional standard deviation in the case of 1d Gaussian, and $\Sigma$ to denote the multidimensional covariance in the case of multidimensional Gaussian.} with $k \ll n_{\mathbf{w}}$, and such that $\mathcal{N}_{\textrm{ws}}$ can be used to approximates the behaviour of $\mathcal{N}_{full}$ in terms of resulting accuracy and uncertainty. 

%\AP{Could be worth it having a graphic explaining the method, if there is time for it.} 
Briefly, we do this by first modelling all the hyperparameters of the $\mathcal{N}_{full}$ distributions through a Gaussian Mixture Model (GMM) (Section \ref{sec:gmm}), and then applying alpha-blending to sample weights from the resulting realisations of the GMM (Section \ref{ellipse}). 
Additionally, 2DGBNN implements several steps informed by best practice in quantisation for deterministic NNs for further reducing the shared number of weights, including outliers detection (Section \ref{outliers}), cluster dimensionality reduction and the merging of similar distributions (Section \ref{sec:gmm}). Finally, we will present the overall algorithm for   2DGBNN in Section \ref{sec:algorithm}.
%However the reduction technique is not applied to the full $\mathbf{w}$ vector but only to common values encountered in the weight-space. Before explaining the details of how the GMM is derived, we describe the weights-classification technique we employ.  

%\AP{Not sure this is needed. Check}

%The parameters are refined to establish an initial approximation of the weight space. This stage sets the stage for the subsequent stochastic network formulation, where the weight distributions are modeled with a mean and variance structure, introducing a principled stochasticity into the network. This stochastic component is critical in capturing the inherent uncertainty within the neural network, a hallmark of Bayesian methods.
%
%The algorithm leverages the trainable GMM, which operates within the weight space to distinguish between inliers and outliers.
%
%In the process of weight classification, the weights of the network are divided into outliers and inliers based on the following factors:
\subsection{Outliers vs. Inliers Classification}
\label{outliers}
The key observation behind the weight-classification stage of our algorithm is that not all the weights of a neural network have an equal impact on the output. Taking inspiration from quantisation techniques for deterministic neural networks \citep{subedar2021quantization}, we, therefore, do not quantise extreme values in the BNN as those are, likely, particularly influential in the final result.  
Specifically, we partition the full weight vector $\textbf{w}$ into two separate vectors, $\textbf{w}_{in}$ and $\textbf{w}_{out}$, and only apply weight-sharing to the former. We do this by using two different criteria.
%In the process of weights, the weights of the network are divided into outliers and inliers based on the following factors:
%\par
\paragraph{Mean Threshold:} Weights associated with a mean with an absolute value greater than a threshold $\tau$ (e.g., $\tau=0.2$) are classified as outliers, i.e.:
\begin{equation*}
w_i \in \mathbf{w} \text{ is an outlier if } \lvert \mu_i \rvert > \tau,
\end{equation*}
A discussion of how we chose parameters like $\tau$ is provided in Appendix \ref{sec:hyperparameters}.

\paragraph{Gradient Threshold:} Weights associated with gradient magnitudes exceeding the threshold that places them within the top 1\% during backpropagation are also categorized as outliers, as their high gradient values likely signify their substantial impact on model performance, i.e.:
\begin{align*}
w_i \in \mathbf{w} \text{ is an outlier if }
\lvert \nabla_{w_i} \rvert \text{ is top 1\% of } \{ \lvert \nabla_{w_j} \rvert \}_{j = 1}^{n_\mathbf{w}}.
\end{align*}
% \subsection{Ellipse Weights:} 
% \AP{This section is long and repetitive, despite the idea being quite simple. Also you talk about cluster assignment, but no clustering has been done so far. Shouldn't it be after 2dgbnn? If you keep it short you'll make space for the pseudocode, which would be good to have in the main text.}
% For all weights classified as inliers in the Bayesian Network, where each weight is represented by a pair of parameters $(\mu, \sigma)$, we compute the probability density function (pdf) values of each weight for all Gaussian distributions (2D). If the pdf value computed for a given cluster is less than 5.991—where 5.991 is the squared Mahalanobis distance between a sample point and the mean, standardized to follow a chi-squared distribution with 2 degrees of freedom—we reassess the cluster assignment.
\begin{algorithm}[!h]
\caption{2DGBNN}\label{alg:pseudocode}
\textbf{Input:} NN architecture $f^{\mathbf{w}}$, training data $\mathcal{D} = \{(\mathbf{X}, \mathbf{y})\}$ – $\tau_w,\tau_d,\tau_g, \tau_v$ algorithm thresholds – BNN prior $p(\mathbf{w})$.\\
\textbf{Output:} Stochastic weight-sharing trained BNN. 
\begin{algorithmic}[1]
\Statex \textbf{Stage 1: Initialise GMM} 
\State  Initialise $\mu$, $\sigma$ according to $p(\mathbf{w})$
\For{each epoch} \Comment{Pre-training}
    \State Sample weights: $\mathbf{w} = \mu + \sigma \odot \epsilon$, $\epsilon \sim \mathcal{N}(0, \mathbf{I})$
    \State Update $\mu$, $\sigma$ by training on $\mathcal{D}$
\EndFor
%\State \textbf{Stage 2: 2D Gaussian and Merging}
    \If{$|w_i| > \tau_w$ \textbf{or} $|\nabla_{w_i}|$ in top $1\%$}  $w_i$ is outlier
    \State  \textbf{else} $w_i$ is inlier \Comment{\S \ref{outliers}}
    \EndIf
\State Learn GMM on inlier params $\Theta_{in}$ (Equation \eqref{eq:gmm_learning})
\Statex \textbf{Stage 2: Refine GMM}
\For{each inlier weight $w_i$} 
    \State Perform \S\ref{ellipse} check on Mahalanobis distance
    \If{Outside 95th percentile}
        \State $w_i$ is assigned to multiple clusters.
        \Else{} $w_i$ is assigned only to the closest Gaussian.
    \EndIf
\EndFor
\State Apply alpha-blending for ellipse points (Eq. \ref{eq:alpha})
\Repeat
    \For{each pair $(\mathcal{N}_1, \mathcal{N}_2)$ in GMM}
        \If{$W(\mathcal{N}_1, \mathcal{N}_2) < \tau_d$, $\Delta_g < \tau_g$, $\Delta_v < \tau_v$}
            \State Merge $\mathcal{N}_1$, $\mathcal{N}_2$ using Eqs.\ \eqref{eq:merge_mean}, \eqref{eq:merge_var}
        \EndIf
    \EndFor
\Until{No more Gaussians can be merged}

\Statex \textbf{Stage 3: Final BNN Training}
\For{each epoch}
    \For{each weight $w_i$}
        \If{$w_i$ is inlier}
            \State Sample $w_i \sim \sum_{k=1}^{K} \pi_k \mathcal{N}(\mu_k, \Sigma_k)$
        \Else
            \State Use $w_i\sim \mathcal{N}(\mu_{w_i}, \sigma^2_{w_i})$
        \EndIf
    \EndFor
    \State Minimising step for $\hat{\mathcal{L}}(\mathcal{D},q)$ of Eq. \eqref{eq:variational_loss}
\EndFor
\end{algorithmic}
\end{algorithm}
\subsection{2DGBNN training}\label{sec:gmm}
Given the set of distributions of the inlier weights, which we denote as $\mathcal{N}_{in}$, we proceed by clustering their means and variances on the 2-dimensional $\mu$-$\sigma$ plane. We do this by learning a Gaussian Mixture Model (GMM) of the form:
%\begin{align*}
 $   p( (\mu,\sigma) ) = \sum_{k=1}^K \pi_k \mathcal{N}\left((\mu,\sigma) | \mu_k, \Sigma_k\right)$
%\end{align*}
over the set of points $\Theta_{in}  =\{(\mu_{w_i},\sigma_{w_i})\}_{i=1}^{n_{\mathbf{w}_{in}}}$. Due to the large volume of points involved in the learning of the GMM (typically, millions of weights), we rely on mini-batch learning for GMMs \citep{li2014efficient}. 
This is achieved by sampling random mini-batches $\mathcal{B} \subset \Theta_{in}$, and iteratively minimising the log-likelihood over the mini-batch:
\begin{align}\label{eq:gmm_learning}
\min \sum_{(\mu_i,\sigma_i) \in \mathcal{B}} \log \left( \sum_{k=1}^{K} \pi_k \mathcal{N}((\mu_i,\sigma_i) \mid \mu_k, \Sigma_k) \right).
\end{align}
% Crucially, mini-batch learning for GMMs can be performed straightforwardly by means of gradient descent, therefore the learning of the GMMs can be seamlessly integrated with the variational approximation of the Bayesian posterior, allowing us to perform stochastic weight-sharing at every step of the learning process. 

After the Initial GMM learning, we perform two further reduction steps based on the number of points around each Gaussian, and on the distance between pairs of Gaussians. 

\paragraph{Cluster Size Reductions:} During the initial Gaussian clustering stage, clusters associated with fewer than 30 weights are identified. Weights within these small clusters are treated as outliers due to their lack of representation within the broader weight distribution.
Overall, we empirically find that approximately $1.8\%$ of the total weights of a neural network are generally allocated as outliers.

\paragraph{Merging Gaussians:} We merge together Gaussian distributions that are very close to each other. We do this by relying on the distance between Gaussians and their gradients. Specifically, we compute the Wasserstein-2 distance between pairs of distributions as \citep{jacobs2023memory}: 
\begin{equation}
\begin{split}
    W_2(\mathcal{N}(\mu_i, \Sigma_i), \mathcal{N}(\mu_j, \Sigma_j))^2 
    = ||\mu_i - \mu_j||_2^2 \\
    \quad + \text{Tr} \left( \Sigma_i + \Sigma_j - 2(\Sigma_i^{1/2} \Sigma_j \Sigma_i^{1/2})^{1/2} \right)
\end{split}
\end{equation}
If the distance between two Gaussians is less than a given threshold $\gamma$, then we inspect the gradient of the network in the weight associated to the cluster centroid and its variance. If those are smaller than two given threshold $\sigma$ and $\alpha$ then we proceed by merging the two Gaussians into one.\footnote{Details about the thresholds we use in our experiments can be found in the Appendix.}

\begin{table*}[!h]
\centering
\caption{Comparison of 2DGBNN and competitive techniques superscripts indicate matching architectures) on \textbf{ImageNet1k} dataset. We also provide the number of outliers, ellipses, and Gaussians derived by our method.}
\renewcommand{\arraystretch}{1.6}
\resizebox{\textwidth}{!}{

\begin{tabular}{@{}ccccccccc@{}}
\toprule
\textbf{Architecture} & Method & 
\textbf{Accuracy $\uparrow$} & 
\textbf{NLL} $\downarrow$ & 
\textbf{ECE} $\downarrow$ & 
\textbf{\#Outliers} & \textbf{\#Ellipses} & \textbf{\#Gaussians}&\textbf{\parbox{4cm}{\centering \#Parameters(M) $\downarrow$ / Compression Ratio(\%) $\uparrow$}}
\\
\hline
\multirow{2}{*}{ResNet-18} & Mutual BNN \citep{pham2024model} & {67.7} & 1.327 &  0.1300 & -&-&-& 23.4M / -\hspace{1em} \\
% \multicolumn{4}{c}{23.4M} 
 & \textbf{2DGBNN(ours)} &  {\textbf{68.1}} & \textbf{1.253} &  \textbf{0.019} & 23013 &10885&2217& \textbf{0.038M} / \textbf{99\%} \\
\cdashline{2-9}
\multirow{5}{*}{ResNet-50}& Deep Ensembles \citep{lakshminarayanan2017simple} & 77.5 & 0.877 & 0.0305&-&-&-& 146.7M / - \hspace{1em}  \\
& Rank-1 BNN\citep{dusenberry2020efficient} & 77.3 & 0.886 &0.0166&-&-&-& 26.0M / - \hspace{1em} \\
& ATMC (30 samples) \citep{heek2019bayesian} &  77.5 & 0.883 & - &- &-&- &768.0M / - \hspace{1em}\hspace{1em}\hspace{1em}\\

& MCMC (9 samples) BNN \citep{zhang2019cyclical} &   77.1 & 0.888 &  - & - & - & - & 230.4M / -  \hspace{1em}\hspace{1em}  \\
&\textbf{2DGBNN(ours)}& 75.1 &0.961 &  0.029 &37172& 56873&3250&\textbf{0.101M}/ \textbf{99\%} \hspace{1em}\hspace{1em}\\
\cdashline{2-9}
% \midrule
% \midrule

% \hline

\textbf{ResNet-101} & \multirow{2}{*}{\textbf{2DGBNN(ours)}} &   75.50 & 0.969 &  \textbf{0.023} & 53641&4311&2464&\textbf{0.063M} / 99\%\hspace{1em} \\
% \cdashline{2-9} % Dashed line from column 2 to 3
% & \multicolumn{1}{|c}{VIT-B-16} &   	
% 81.07 & 0.856 & 0.056 &-&-&-&86M  \\
\textbf{VIT-B-16} && 76.01 &  0.901 &  0.064& 9765&338329&5440&\textbf{0.359M }/ 98\%\hspace{1em}\\

\hline
\end{tabular}
}

\label{tab:imagenet1k}
\end{table*}

% 538

The merger is executed using the following equations \citep{agueh2011barycenters, takatsu2011wasserstein}:
\begin{equation}\label{eq:merge_mean}
\mu_{\text{merged}} = \frac{\mu_1 + \mu_2}{2}
\end{equation}
\begin{equation}\label{eq:merge_var}
\begin{split}
    \Sigma_{\text{merged}} = \frac{\Sigma_1 + \Sigma_2}{2} 
    &+ \frac{1}{8}(\mu_1 - \mu_2)(\mu_1 - \mu_2)^T \\
    &+ \frac{1}{2} \left( \Sigma_1^{1/2} \Sigma_2 \Sigma_1^{1/2} \right)^{1/2}
\end{split}
\end{equation}
%This formula refines the merger of Gaussian components by averaging their centroids and covariance matrices and incorporating a correction factor, \(\frac{1}{8}(\mu_1 - \mu_2)(\mu_1 - \mu_2)^T\), for the Euclidean distance between centroids. Additionally, it integrates the geometric mean of the covariance matrices, \((\Sigma_1^{1/2}\Sigma_2\Sigma_1^{1/2})^{1/2}\), to precisely reflect the statistical properties of the distributions.
%
That ensures that the newly formed Gaussian component accurately reflects the collective distribution characteristics of the initial components while maintaining minimal internal variation.

\subsection{$\alpha$-blending (Multi-Clusters) for Weights}
\label{ellipse}
Before the final sampling step, we reassess inlier weights $\mathbf{w}_{in}$ by computing their squared Mahalanobis distances to cluster means using, i.e.,
\(
D^2 = (w_i - \mu_i)^\top \Sigma_k^{-1} (w_i - \mu_i)
\).
%where $D^2$ follows a chi-squared distribution with 2 degrees of freedom. 
If a weight's $D^2$ exceeds 5.991,\footnote{Corresponding to the 95th percentile of the $\chi^2_2$ distribution which models Mahlanobis distance of multidimensional Gaussians.} we reassess its cluster assignment:
%The final step of our methodology is the actual sampling of the weights from the 2dGMM built over the $\mu-\sigma$ space. 
We do this by relying on $\alpha$-blending \citep{mildenhall2021nerf}.

Specifically, for each the $w_i \in \mathbf{w}_{in}$ we compute the subset of GMM's component $\mathcal{N}(\mu_k,\Sigma_k)$, for $k=1,\ldots,n_i$ such that the above condition on the $D^2$ is met. We then sample the final value of the weight by the resulting distributions:
\begin{equation}\label{eq:alpha}
p(w_i) = \sum_{k=1}^{n_i} \alpha_k \mathcal{N}( \mu_k, \Sigma_k)
\end{equation}
where $\alpha_k$ is the mixing coefficient, computed as the pdf of $(\mu_i,\sigma_i^2)$ according to $\mathcal{N}( \mu_k, \Sigma_k)$.

\begin{figure*}[t]
    \centering
    \includegraphics[width=0.92\textwidth]{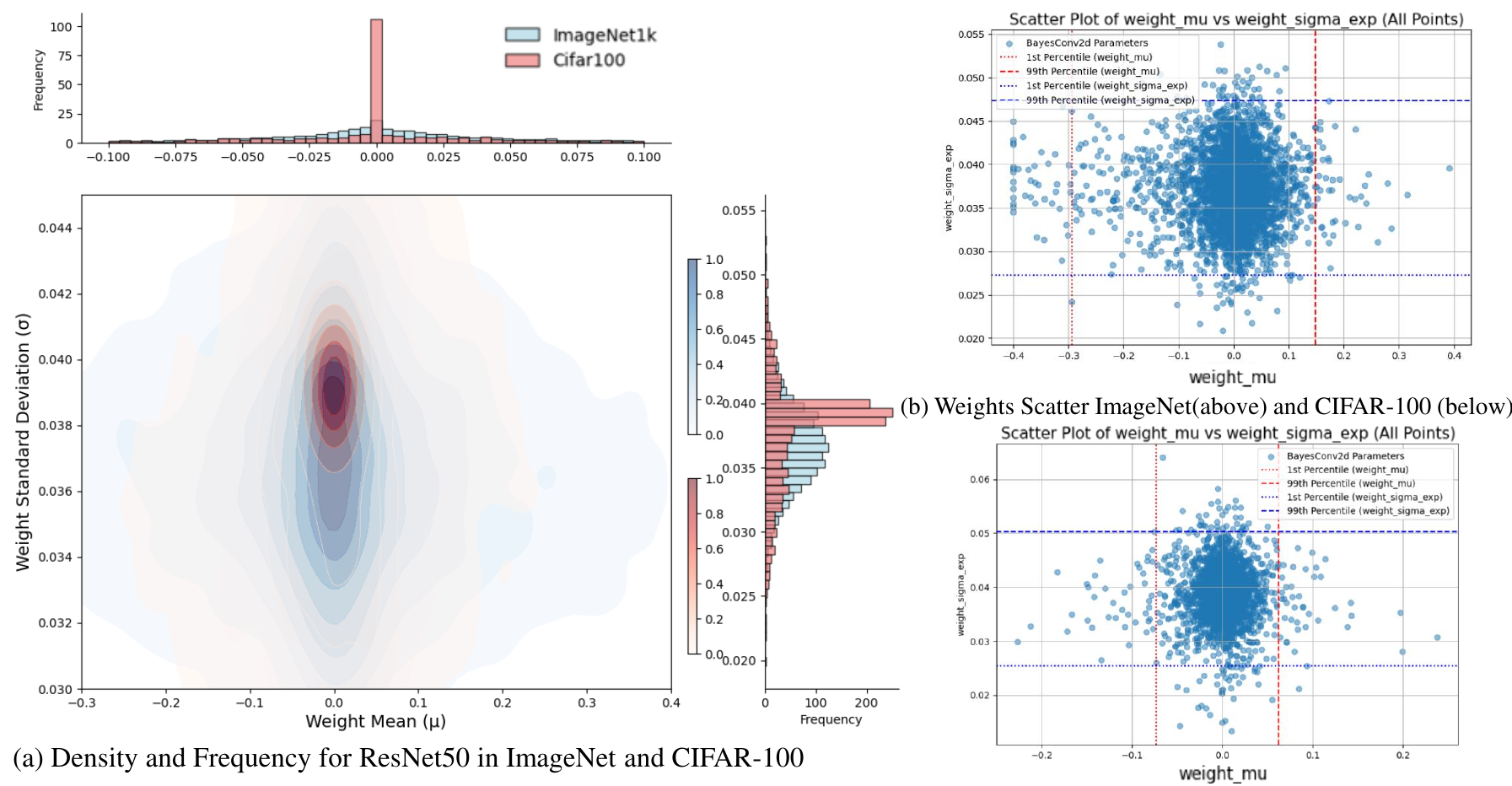}
    \caption{Weight distribution for the BNN prior to stochastic-sharing. Panel (a) shows the density plot for the second convolutional layer of ResNet-50 when trained on ImageNet (in blue) and CIFAR-100 (in red). Panel (b) shows the corresponding scatter plots, including lines for the 1\% and 99\%.}
%    \vspace{-0.3cm}
    \label{fig:den}
\end{figure*}

\subsubsection{Combined Variational Formulation}

Finally, we observe that our stochastic weight-sharing technique can be seamlessly integrated within the ELBO formulation for VI training \citep{nowlan2018simplifying, zhang2018advances}. 
Formally, we assume that $\mathbf{w}_{out}$ and $\mathbf{w}_{in}$ are vectors of pairwise independent weights,\footnote{This is true for the variational distribution but it is an approximation for the true posterior.} which allow us to bound the variational objective as it follows:  
\begin{align}
&\mathcal{L}(\mathcal{D},q) =\nonumber \\&  \mathbb{E}_{q(\mathbf{w})}\left[ \log p(\mathbf{y} \mid \mathbf{X}, \mathbf{w}) \right] - \text{KL}\left( q(\mathbf{w}) \,\Vert\, p(\mathbf{w}) \right)  = \label{eq:equality}\\ 
&\mathbb{E}_{q(\mathbf{w})}\left[ \log p(\mathbf{y} \mid \mathbf{X}, \mathbf{w}) \right] - \text{KL}\left( q(\mathbf{w}_{in}) \,\Vert\, p(\mathbf{w}_{in}) \right)  - \nonumber\\
&\text{KL}\left( q(\mathbf{w}_{out}) \,\Vert\, p(\mathbf{w}_{out}) \right)  \approx \mathbb{E}_{q(\mathbf{w})}\left[ \log p(\mathbf{y} \mid \mathbf{X}, \mathbf{w}) \right] \label{eq:approximation} \\ & -\text{KL}\left( \sum \pi_k \mathcal{N}_k \,\Vert\, p(\mathbf{w}_{in}) \right)  - \text{KL}\left( q(\mathbf{w}_{out}) \,\Vert\, p(\mathbf{w}_{out}) \right) \nonumber \\ 
&\geq \mathbb{E}_{q(\mathbf{w})}\left[ \log p(\mathbf{y} \mid \mathbf{X}, \mathbf{w}) \right] - \sum_k \pi_k \underbrace{\text{KL}\left( \mathcal{N}_k \,\Vert\, p(\mathbf{w}_{in}) \right)}_{\text{GMM KL divergence}} \nonumber - \\
& \underbrace{\text{KL}\left( q(\mathbf{w}_{out}) \,\Vert\, p(\mathbf{w}_{out}) \right)}_{\text{Outliers KL divergence}} := \hat{\mathcal{L}}(\mathcal{D},q), \label{eq:variational_loss}
%& - \sum_{k=1}^{K} \pi_k \underbrace{\text{KL}\left( q(\mathbf{w}_k) \,\Vert\, \mathcal{N}(\mu_k, \Sigma_k) \right)}_{\text{GMM KL divergence}} \\
%& - \underbrace{\text{KL}\left( q(\mathbf{w}_{\text{Outliers}}) \,\Vert\, p(\mathbf{w}_{\text{Outliers}}) \right)}_{\text{Outlier KL divergence}}
\end{align}
where the equality in Equation \eqref{eq:equality} is due to the pairwise independence assumption between components of $\mathbf{w}_{in}$ and $\mathbf{w}_{out}$, the approximation of Equation \eqref{eq:approximation} is due to the GMM approximation of the inlier weights, and the final inequality is due to the convexity of the KL divergence. Notice that the resulting value loss function, $\hat{\mathcal{L}}$ is an upper bound on the original loss ${\mathcal{L}}$ so that its minimisation by means of gradient descent guarantees the improvement of the latter.

\subsection{Overall Methodology}\label{sec:algorithm}
The overall methodology is presented in pseudocode form in Algorithm \ref{alg:pseudocode}. 
2DGBNN combines its component parts in three stages. In the first stage, the BNN is initialised with the given prior, a pre-training step is performed and the resulting BNN is used to initialise the GMM clustering.
In stage 2, the initial GMM clustering obtained is refined by merging close Gaussians, and by performing $\alpha$-blending. Finally the BNN is trained by optimising the variational objective on a combination of full weight distributions (for the outliers) and GMM-based weight-sharing (for the inliers).
%We aims to approximate the full set of weight distributions $\mathcal{N}{\text{full}}$ by \textit{a significantly reduced set of shared Gaussian distributions} $\mathcal{N}_{\text{ws}}$, where $k \ll n_{\mathbf{w}}$. We begin by classifying weights into outliers and inliers based on mean and gradient thresholds, preserving the distributions of outliers due to their significant impact on the model. For inlier weights, we cluster their mean and variance pairs in the 2D $\mu$-$\sigma$ space using a Gaussian Mixture Model (GMM), trained via mini-batch learning to handle large datasets. We further reduce the number of clusters by merging similar Gaussians based on the Wasserstein-2 distance and additional gradient criteria. Ellipse weights are assigned to multiple clusters, and employ $\alpha$-blending to sample from the blended distributions. The outlier weights retain their original distributions to maintain the model's performance and uncertainty estimates. This stochastic weight-sharing approach significantly reduces computational complexity without compromising accuracy.

\begin{table*}[h]
\centering
\caption{Comparison of 2DGBNN and competitive techniques (superscripts indicate matching architectures) on \textbf{CIFAR-100} dataset. We also provide the number of outliers, ellipses, and Gaussians derived by our method.}
\renewcommand{\arraystretch}{1.6}
\resizebox{\textwidth}{!}{
\begin{tabular}{@{}ccccccccc@{}}
\toprule
\textbf{Architecture} & Method & 
\textbf{Accuracy $\uparrow$} & 
\textbf{NLL} $\downarrow$ & 
\textbf{ECE} $\downarrow$ & 
\textbf{\#Outliers} & \textbf{\#Ellipses} & \textbf{\#Gaussians}&\textbf{\parbox{4cm}{\centering \#Parameters(M) $\downarrow$ / Compression Ratio(\%) $\uparrow$}}
% \multirow{8}{*}{\textbf{\textbf{\# Parameters}}}
\\
\midrule
\multirow{4}{*}{ResNet-18} &F-SGVB-LRT \citep{nguyen2024flat}&70.1 &  1.121& 0.036&-&-&-& 23.4M / -\hspace{1em} \\
& SSVI \citep{li2024training}& 75.8  & - & 0.001 &-&-&-& 2.32M / 90\%  \\
& mBCNN \citep{kong2023masked}& 73.7 & 1.004 & 0.002 &-&-&-& 2.86M / 87.8\%  \\

& \textbf{2DGBNN(ours)} &  74.7 & 1.053 &  0.038 & 14624 &260&2387& \textbf{0.019M} / \textbf{99}\% \\

\cdashline{2-9}
\multirow{3}{*}{WRN-28-10} & Deep Ensembles \citep{lakshminarayanan2017simple}& 82.7 & 0.666 & 0.021&-&-&-&  146M / -  \\
&Rank-1 BNN \citep{dusenberry2020efficient} & 82.4 & 0.689 &0.012& -&-&-& 36.6M / -\\
 & LP-BNN \citep{franchi2023encoding} &  79.3 & - & 0.0702  &-&-&-& 26.8M / 63\% \\
&  \textbf{2DGBNN(ours)} &   80.5 & 0.798 & 0.0432 & 40354&341&2390&\textbf{0.045M} / \textbf{99\%}\\
\cdashline{2-9}
% \midrule
% \midrule
\multirow{2}{*}{ResNet-50} & ABNN \citep{franchi2024make} &  74.20 & 0.828&  4.5& -&-&-& 54.2M / - \\
&  \textbf{2DGBNN(ours)} &   78.1 & 0.986 &  0.107 &247591& 330&1980&\textbf{0.251M} / \textbf{99\%}  \\
% \hline
\cdashline{2-9}
% & \multicolumn{1}{|c}{ResNet-101} &   	
% 80.0 & 0.849 & 0.095 & -&-&-&44.5M  \\
ResNet-101&  \textbf{2DGBNN(ours)}  &  78.4 & 0.834 &  0.066 & 45240&348&3199&\textbf{0.052M} / \textbf{99\%} \\
% \cdashline{2-9} % Dashed line from column 2 to 3
% & \multicolumn{1}{|c}{WRN-28-10} &   	
% 81.4 & 0.766 & 0.0456 &-&-&-&53.6M  \\
% WRN-28-10 & & 80.5 & 0.798 & 0.0432 & 40354&341&2390&\textbf{0.045M}  \\
\bottomrule
\end{tabular}
}
\label{tab:cifar100}
\end{table*}

\section{EXPERIMENTS}\label{sec:experiments}

To validate the effectiveness and scalability of 2DGBNN, we conduct comprehensive experiments using various NN architectures on benchmark image classification datasets. This section details the datasets, models, experimental setup, results, and analysis of our findings.
Specifically, we evaluate our method on four common image classification benchmarks: MNIST, CIFAR-10, CIFAR-100, and ImageNet1k; as well as four widely used NN architectures: ResNet-18, ResNet-50 and ResNet-101 and a Vision Transformer (ViT). The hyperparameters used and the details of the training are given in Appendix. Throughout this section, we compare our technique against the results obtained by Deep Ensembles \citep{lakshminarayanan2017simple}, Rank-1 BNN \citep{dusenberry2020efficient}, MCMC BNN \citep{zhang2019cyclical}, ATMC \citep{heek2019bayesian}, Mutual BNN \citep{pham2024model}, F-SGVB-LRT \cite{nguyen2024flat}, ABNN \citep{franchi2024make}, LP-BNN \citep{franchi2023encoding}, IR \citep{kim2023inverse}, SSVI \citep{li2024training} and mBCNN \citep{kong2023masked}. 
We evaluate the resulting models in terms of Accuracy, Negative Log-Likelihood (NLL, which measures the model's uncertainty in its predictions), and Expected Calibration Error (ECE, which measures the calibration of predicted probabilities~\citep{guo2017calibration}).
Each experiment is conducted three times with different random seeds, and we report the average results.  We conduct experiments with all the aforementioned comparison models, totaling 13 2DGBNN experiments, which include the base models of all comparison methods. Additionally, we perform two quantisation comparison experiments (Section \ref{quantisation}) and an ablation study (Section \ref{sec:ablation}). %And also we conduct an ablation studies to demonstrate the effectiveness of our module.

% \AP{You should mention some numbers to give an idea about how many experiments you did. For example, you could say how many models you ended up training using 2DGBNN for the comparisons (counting all the random seeds)}
%

\begin{table*}[h]
\centering
%\caption{Comparison of different methods in \textbf{CIFAR-10} dataset, in here, model compression methods included. The first five rows present the methods we compared, with superscripts in the upper-right corner indicating literature references: IR\textsuperscript{0} \citep{kim2023inverse}, F-SGVB-LRT\textsuperscript{1} \citep{nguyen2024flat}, Rank-1 BNN\textsuperscript{1}\citep{dusenberry2020efficient}, ABNN\textsuperscript{3} \citep{franchi2024make}. In the 2DGBNN section, \textbf{}\textit{\textbf{represents our method, while entries without  denote the counterpart deterministic models}}. The superscripts here denote comparisons under the same framework; for example, ABNN\textsuperscript{3} \citep{franchi2024make} and our ResNet-50\textsuperscript{3} are compared. The columns "Outliers," "Ellipses," and "Gaussians" represent counts of parameters.}
\caption{Comparison of 2DGBNN and competitive techniques (superscripts indicate matching architectures) on \textbf{CIFAR-10} dataset. We also provide the number of outliers, ellipses, and Gaussians derived by our method.}
\renewcommand{\arraystretch}{1.5}
\resizebox{\textwidth}{!}{
\begin{tabular}{@{}ccccccccc@{}}
\toprule
\textbf{Architecture} & Method & 
\textbf{Accuracy $\uparrow$} & 
\textbf{NLL} $\downarrow$ & 
\textbf{ECE} $\downarrow$ & 
\textbf{\#Outliers} & \textbf{\#Ellipses} & \textbf{\#Gaussians}&\textbf{\parbox{4cm}{\centering \#Parameters(M) $\downarrow$ / Compression Ratio(\%) $\uparrow$}}
% \multirow{8}{*}{\textbf{\textbf{\# Parameters}}}
\\
%\hline
\midrule
% \midrule
\multirow{4}{*}{ResNet-18} & F-SGVB-LRT \citep{nguyen2024flat} & 90.31 & 0.262& 0.014&-&-&-&23.4M / - \hspace{1em} \\
& SSVI \citep{li2024training}& 93.74  & - & 0.006 &-&-&-&  1.17M / 95\%  \\
& mBCNN \citep{kong2023masked}& 93.20 &0.220 & 0.008 &-&-&-& 0.93M / 96\%  \\
& \textbf{2DGBNN(ours)} & 91.72&  0.305 &  0.019 & 123310 &67&1569& \textbf{0.018M} / \textbf{99}\%\hspace{1em} \\
\cdashline{2-9}

\multirow{4}{*}{WRN-28-10} & Deep Ensembles \citep{lakshminarayanan2017simple}& 96.2 & 0.143 & 0.020&-&-&-& 146M / -\hspace{1em}   \\
&Rank-1 BNN \citep{dusenberry2020efficient} & 96.3 & 0.128 &0.008& -&-&-& 36.6M / 50.8\%\\
 & LP-BNN \citep{franchi2023encoding} &  95.0 & - &  0.009  &-&-&-& 26.8M / 63\% \\
&  \textbf{2DGBNN(ours)} &   95.2 & 0.142 & 0.012 & 39395&365977&3950&\textbf{0.413M} / \textbf{99\%} \\
\cdashline{2-9}
\multirow{2}{*}{ResNet-50} & ABNN \citep{franchi2024make} &  95.01 & 0.160&  1.0& -&-&-& 54.2M / 25\% \\
&  \textbf{2DGBNN(ours)} &   93.84 & 0.223 & 0.012 &129640& \textbf{0}&1628&\textbf{0.132M} / \textbf{99\%}  \\
\cdashline{2-9}
% ResNet-20 &\textbf{2DGBNN(ours)} &  91.04 & 0.303 & 0.037  & 14624 &260&2387& \textbf{0.020M} / \textbf{99\%}  \\
ResNet-101& \textbf{2DGBNN(ours)}& 92.78 & 0.270 & 0.015 & 45240&348&3199&\textbf{0.052M} / \textbf{99\%}  \\

\bottomrule
\end{tabular}
}

\label{tab:cifar10}
\end{table*}

\subsection{Performance Evaluation}
%\subsubsection{Performance Evaluations}
The results obtained with 2DGBNN and those of the state-of-the-art are listed in Tables \ref{tab:imagenet1k}, \ref{tab:cifar100} and \ref{tab:cifar10} for ImageNet1k, CIFAR-100 and CIFAR-10 respectively along with details on the computations performed by 2DGBNN.

In the case of ImageNet1k (Table~\ref{tab:imagenet1k}) the comparison is performed against Deep Ensembles  \citep{lakshminarayanan2017simple}, Rank-1 BNN \citep{dusenberry2020efficient}, MCMC BNN with 9 samples \citep{zhang2019cyclical}, ATMC \citep{heek2019bayesian}, and Mutual BNN  \citep{pham2024model}.
% demonstrates the performance of our algorithm across different models in ImageNet1k. 
%
We observe that, in all cases, our method successfully reduces the number of parameters by 3 or 4 orders of magnitudes.%\footnote{Notice that the overall storage differential is however smaller as our method requires a uint16 index for each inlier weight, yet still compresses approximately 75\% compared to competitive techniques.} 
While the accuracy is reduced by around $2\%$ in the ResNet-50 case, we do obtain comparable uncertainty estimation as evaluated by NLL and ECE.

Similar results we obtain on the CIFAR-100 dataset (Table \ref{tab:cifar100}), comparing against Deep Ensembles  \citep{lakshminarayanan2017simple}, Rank-1 BNN\citep{dusenberry2020efficient}, F-SGVB-LRT \citep{nguyen2024flat} and ABNN \citep{franchi2024make}.
Our method achieves a substantial reduction in model parameters while maintaining or improving performance compared to other methods at the price of approximately $2\%$ when compared to Deep Ensembles and Rank-1 BNN. For instance, for ResNet-18 we use only 0.019M parameters, which is drastically lower than the 23.4M parameters used by the F-SGVB-LRT model, yet we achieve competitive accuracy. Finally, Table \ref{tab:cifar10} lists analogous results in the context of CIFAR-10,
 comparing against IR \citep{kim2023inverse}, F-SGVB-LRT \citep{nguyen2024flat}, ABNN \citep{franchi2024make} and LP-BNN \citep{franchi2023encoding}. 
 
Additionally to the architectures used for comparisons, the tables report results for ResNet-101 and ViT. In these architectures too, 2DGBNN is able to reduce the number of parameters while obtaining accuracy and uncertainty metrics on par with that of state-of-the-art techniques across the remaining architectures.
Interestingly, observing the training results obtained we notice how the distribution of weights in models trained on smaller datasets (CIFAR-100 and CIFAR-10) tends to cluster near zero, as depicted in Figure~\ref{fig:den} (b) (down). Conversely, in ImageNet1k  the weight distribution is broader, as can be seen from Figure~\ref{fig:den} (a)) comparing empirical distributions obtained on CIFAR-100 and ImageNet1k. % where the histogram shows that weights from CIFAR-100 mainly cluster around zero, whereas the distribution for ImageNet1k is more evenly spread across various values.
%
%Additionally, from the density of Figure.~\ref{fig:den} (a) the ($\sigma$) of the weights in models trained on ImageNet1k is lower compared to those trained on CIFAR-100. This indicates greater stability in training Bayesian Neural Networks (BNNs) on the larger ImageNet1k dataset due to its reduced weight variance.
%
Notice how this translates to, for example, the ResNet-50 model trained on ImageNet1k dataset to have a significantly greater number of Gaussian and Ellipse weights than when trained on the CIFAR-100 dataset.

\begin{table*}[!h]
\centering
\caption{Comparison against the quantisation technique of \citet{subedar2021quantization} on \textbf{CIFAR-10} and \textbf{MNIST}.}
\renewcommand{\arraystretch}{1.7}
\resizebox{\textwidth}{!}{
\begin{tabular}{@{}cccccccccc@{}}
\toprule
\multirow{2}{*}{\textbf{Datasets}} & \multirow{2}{*}{\textbf{Algorithm}} &\multirow{2}{*}{\textbf{Quantisation technique}}&
\multirow{2}{*}{\textbf{Accuracy $\uparrow$}} & 
\multirow{2}{*}{\textbf{NLL} $\downarrow$} & 
\multirow{2}{*}{\textbf{ECE} $\downarrow$} & 
\multicolumn{4}{c}{\textbf{\#Parameters (MB: Megabyte)}}
% \multirow{8}{*}{\textbf{\textbf{\# Parameters}}}
\\
& && & & & \textbf{\#Outliers} & \textbf{\#Ellipses} & \textbf{\#Gaussians}&\textbf{\#Parameters}
\\
\midrule
%\hline
%\hline
\multirow{6}{*}{\textbf{CIFAR-10}} 
&\multirow{3}{*}{\shortstack{\textbf{BNNs Quantization} \\ \citep{subedar2021quantization}}} 
& ResNet-20 (INT8 SIGMA4) &  90.92  &  0.266  &  1.778&-&-&-& \textbf{0.87 MB} \\
& & ResNet-20 (INT8 SIGMA2) &  90.85  & 0.273 &  2.547  &-&-&-& \textbf{0.72 MB} \\
& & ResNet-20 (INT8 SIGMA1) &  90.96  &0.266& 0.711   &-&-&-& \textbf{0.54 MB} \\
\cdashline{3-10}

% &\multirow{4}{*}{\textbf{2DGBNN}}& & & & & \textbf{\#Outliers} & \textbf{\#Ellipses} & \textbf{\#Gaussians}&\textbf{\#Parameters} \\
% & Det &  91.04  &  \textbf{0.303 } &  0.037&  14624 &260&2387& \textbf{0.375 MB}  \\
&\multirow{2}{*}{\textbf{2DGBNN}}& ResNet-20(without pretrained) &  90.91  &  0.303 &  \textbf{0.040}&  74181 &3634&142& \textbf{0.644MB (1.62MB)}  \\
&& ResNet-20(with pretrained) &  \textbf{91.04} & 0.303 & \textbf{0.037}  & 14624 &260&2387& \textbf{0.020M} / \textbf{(0.71MB)}  \\
\midrule
% \multicolumn{2}{c}{\textbf{MNIST}} & 
% \textbf{Accuracy $\uparrow$} & 
% \textbf{NLL} $\downarrow$ & 
% \textbf{ECE} $\downarrow$ & 
% \multicolumn{4}{c}{\textbf{\#Parameters (MB: Megabyte)}}
% % \multirow{8}{*}{\textbf{\textbf{\# Parameters}}}
% \\

\multirow{5}{*}{\textbf{MNIST}}& \multirow{3}{*}{\shortstack{\textbf{BNNs Quantization} \\ \citep{subedar2021quantization}}}
& ResNet-20 (INT8 SIGMA4) & 99.36  &  0.020 &   0.215&-&-&-& \textbf{0.10 MB} \\
&& ResNet-20 (INT8 SIGMA2)  &  99.32  & 0.024 &  0.277 &-&-&-&\textbf{0.08 MB} \\
&& ResNet-20 (INT8 SIGMA1) &  99.34 &0.027&  0.351  &-&-&-&\textbf{0.06 MB} \\

% &&\multirow{3}{*}{\textbf{2DGBNN}}& ResNet-20(without pretrained)& &&  \textbf{\#Outliers} & \textbf{\#Ellipses} & \textbf{\#Gaussians}&\textbf{\#Parameters} \\

% & Det &  99.51  &  \textbf{ 0.011}  &  \textbf{0.001}&  10345 &1106&219& \textbf{0.147MB (0.541MB)}  \\
\cdashline{3-10}
&\textbf{2DGBNN}& ResNet-20(without pretrained)  &  \textbf{99.52}  &  0.013  & \textbf{ 0.001}&  3206 &581&237& \textbf{0.092MB (0.403MB)}  \\
\bottomrule
\end{tabular}
}

\label{tab:mnist}
\end{table*}

\begin{table*}[h]
\centering
\vspace{0.25cm}
\caption{Ablation Study: Impact of Outliers and Ellipses on CIFAR-10 Using ResNet-20} 
\label{tab:ablation}
\begin{tabular}{lcccccc}
\toprule
\textbf{Configuration} &  \textbf{\#Outliers} & \textbf{\#Ellipses} &\textbf{Accuracy (\%)} $\uparrow$ & \textbf{NLL} $\downarrow$ & \textbf{ECE}$\downarrow$ \\
\midrule
2DGBNN    & \checkmark & \checkmark & 90.92&0.265 &0.040  \\
Without Ellipse & \checkmark & --
  & 90.43& 0.304&0.043 \\
Without Outliers & -- & \checkmark & 90.18&0.308 &0.026 \\
Without Outliers and Ellipse & --  & --  &90.01&0.318 &0.029 \\
\bottomrule
\end{tabular}
\end{table*}

\subsection{Comparison against Quantisation}
\label{quantisation}
We now compare 2DGBNN against quantisation techniques applied to BNNs \citep{subedar2021quantization}. For this purpose, we remove the pretraining stage of 2DGBNN so to mimic the ``vanilla'' BNN training employed by \citet{subedar2021quantization}. We use a Gaussian prior with a mean of 0 and a standard deviation of 0.1. Notice that these experiments are limited to CIFAR-10 and MNIST as the vanilla training of BNNs used in \citet{subedar2021quantization} does not scale to the larger architectures and datasets analysed in the previous section. 

The comparative results are presented in Table \ref{tab:mnist}. Accuracy values are very similar across the board, while our technique obtains significantly better uncertainty metrics, except for NLL in the case CIFAR-10. 

In terms of the model size (here compared in Megabytes), the two techniques compare similarly when it comes to the size of trainable parameters (corresponding to the value reported not in brackets for 2DGBNN), with quantisation having a slight edge when only 1 bit is used for encoding the standard deviation. Notice, however, that in small NNs (like the one here analysed) our techniques incur significant storage overhead in that we need to keep an index (encoded in uint8) that assigns each inlier weight to its cluster. When this value is added (size reported in brackets in the Table) quantisation has a significant advantage over our storage requirements. While techniques such as Huffman coding or multi-level index tables can potentially reduce the size of the index vector by several factors, we leave further investigations to future work, and here notice that despite maintaining full precision on the workings of the BNN, weight-sharing quantisation can already obtain comparable results to int8 quantisation. We notice that the two methods are complimentary, and int8 quantisation can further reduce the storage requirements of the outlier weights and GMMs.

\subsection{Ablation Study}\label{sec:ablation}
Table \ref{tab:ablation} presents an ablation study on CIFAR-10 using ResNet-20 to explore how outliers and ellipses contribute to the performance of 2DGBNN. When both outliers and ellipses are included, the model achieves an accuracy of 90.92\%, with the lowest NLL of 0.265 and ECE of 0.040. However, removing either component significantly impacts performance. Without ellipses, the accuracy drops by 0.49\% to 90.43\%, and excluding outliers reduces it slightly further by 0.25\% to 90.18\%. When both are removed, the accuracy drops by 0.87\% to a low of 90.05\%. 
\section{CONCLUSIONS}
We have presented a stochastic weight-sharing quantisation technique based on GMMs specifically tailored to BNNs. In an extensive empirical evaluation, we have seen how our technique can significantly reduce the effective number of parameters of a BNN while obtaining results on par with state-of-the-art in large datasets and architectures such as ImageNet1k and ViT. 

Future work will explore how to integrate our method into a fully Bayesian framework and the application of further quantisation for the outlier weights.  
We have presented a stochastic weight-sharing quantisation technique based on GMMs specifically tailored to BNNs. In an extensive empirical evaluation, we have seen how our technique can significantly reduce the effective number of parameters of a BNN while obtaining results on par with state-of-the-art in large datasets and architectures such as ImageNet1k and ViT. Future work will explore how to integrate our method into a fully Bayesian framework and the application of further quantisation for the outlier weights.  
\section{ACKNOWLEDGEMENTS}
This publication has emanated from research jointly funded by 
European Union’s Horizon Europe 2021–2027 framework programme, Marie Skłodowska-Curie Actions, Grant Agreement No. 101072456 and
Taighde Éireann – Research Ireland under grant number 13/RC/2094\_2.
\bibliographystyle{apalike} % 选择引用格式
\bibliography{iclr2025_conference}
\newpage
\section*{Checklist}
 \begin{enumerate}

 \item For all models and algorithms presented, check if you include:
 \begin{enumerate}
   \item A clear description of the mathematical setting, assumptions, algorithm, and/or model. Yes
   \item An analysis of the properties and complexity (time, space, sample size) of any algorithm. Yes
   \item (Optional) Anonymized source code, with specification of all dependencies, including external libraries. Yes
 \end{enumerate}

 \item For any theoretical claim, check if you include:
 \begin{enumerate}
   \item Statements of the full set of assumptions of all theoretical results. Yes
   \item Complete proofs of all theoretical results. Not Applicable
   \item Clear explanations of any assumptions. Yes  
 \end{enumerate}

 \item For all figures and tables that present empirical results, check if you include:
 \begin{enumerate}
   \item The code, data, and instructions needed to reproduce the main experimental results (either in the supplemental material or as a URL). Yes
   \item All the training details (e.g., data splits, hyperparameters, how they were chosen). Yes
         \item A clear definition of the specific measure or statistics and error bars (e.g., with respect to the random seed after running experiments multiple times). Yes
         \item A description of the computing infrastructure used. (e.g., type of GPUs, internal cluster, or cloud provider). Yes
 \end{enumerate}

 \item If you are using existing assets (e.g., code, data, models) or curating/releasing new assets, check if you include:
 \begin{enumerate}
   \item Citations of the creator If your work uses existing assets. Yes
   \item The license information of the assets, if applicable. Yes
   \item New assets either in the supplemental material or as a URL, if applicable. Not Applicable
   \item Information about consent from data providers/curators. Not Applicable
   \item Discussion of sensible content if applicable, e.g., personally identifiable information or offensive content.Not Applicable
 \end{enumerate}

 \item If you used crowdsourcing or conducted research with human subjects, check if you include:
 \begin{enumerate}
   \item The full text of instructions given to participants and screenshots. Not Applicable
   \item Descriptions of potential participant risks, with links to Institutional Review Board (IRB) approvals if applicable. Not Applicable
   \item The estimated hourly wage paid to participants and the total amount spent on participant compensation. Not Applicable
 \end{enumerate}

 \end{enumerate}
\onecolumn  % 切换到单栏模式
\clearpage
\appendix
\newpage
\aistatstitle{Stochastic Weight Sharing for Bayesian Neural Networks \\ Supplementary Materials}
\vspace{-5em}
% \setcounter{table}{0}
% \setcounter{figure}{0}
%\renewcommand{\thefigure}{A.\arabic{figure}}
% \renewcommand{\thefigure}{\thesection.\arabic{figure}}
% \renewcommand{\thetable}{\thesection.\arabic{table}}
% \counterwithin{figure}{section}
% \counterwithin{table}{section}
% 重置表格计数器
\section{Posterior Visualisation}
\begin{figure}[htbp]
\centering
\includegraphics[width=1\linewidth]{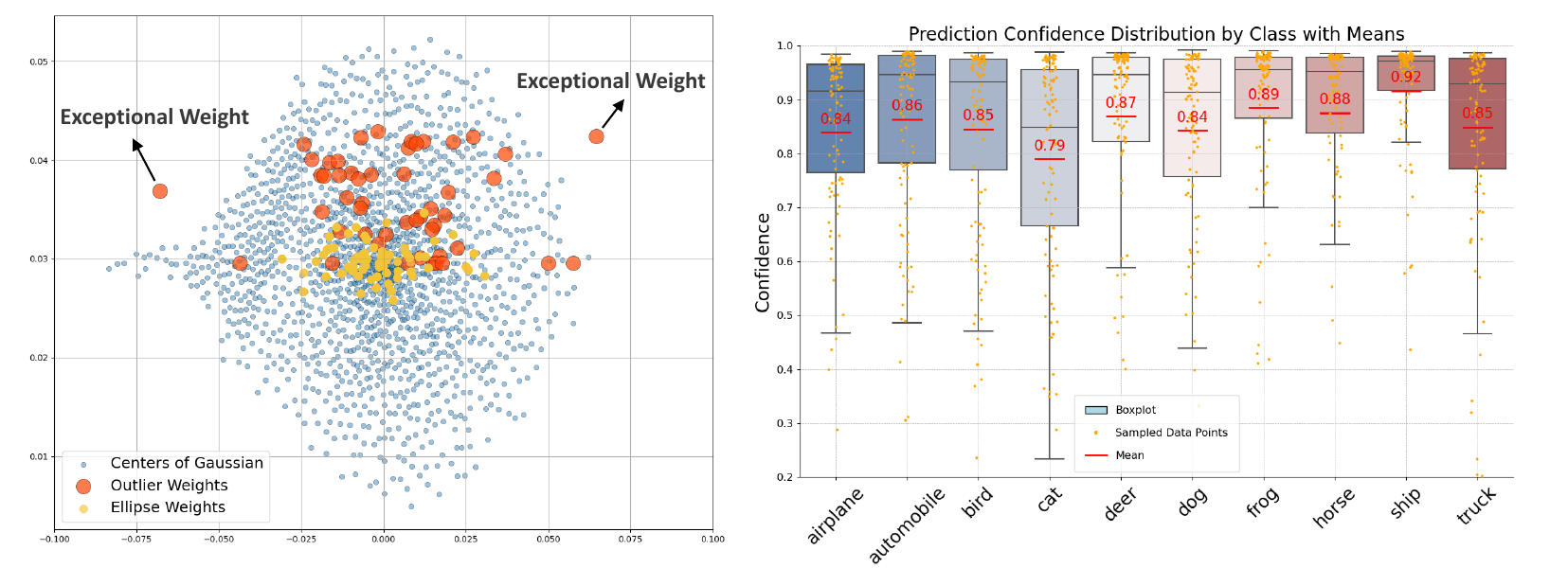}
\caption{Left: Posterior distribution of weights in ResNet-18 with CIFAR-10 using  2D Gaussian Bayesian Neural Network (2DGBNN), depicting Gaussian centers (blue), Ellipse weights (yellow), and Outlier weights (red). Exceptional weights (outliers). Right: Confidence distribution of predictions on the CIFAR-10 dataset.  Mean confidence values are shown above each boxplot.}
\label{fig:posteriorandoutput}
\end{figure}

The posterior distribution of weights provides direct insight into the behaviour of 2D Gaussian Bayesian Neural Network (2DGBNN). To present this, we visualise the posterior distribution of ResNet-18 trained on the CIFAR-10 dataset in Figure~\ref{fig:posteriorandoutput} (left plot).
In the figure, red points represent outlier weights that retain their individual values instead of participating in weight sharing, while yellow points denote ellipse weights shared through alpha-blending. Blue points indicate the centres of the Gaussian distributions after training (note that for simplicity of visualisation, we show only 20 outlier weights and 50 ellipse weights).

\begin{figure}[htbp]
\centering
\includegraphics[width=0.9\linewidth]{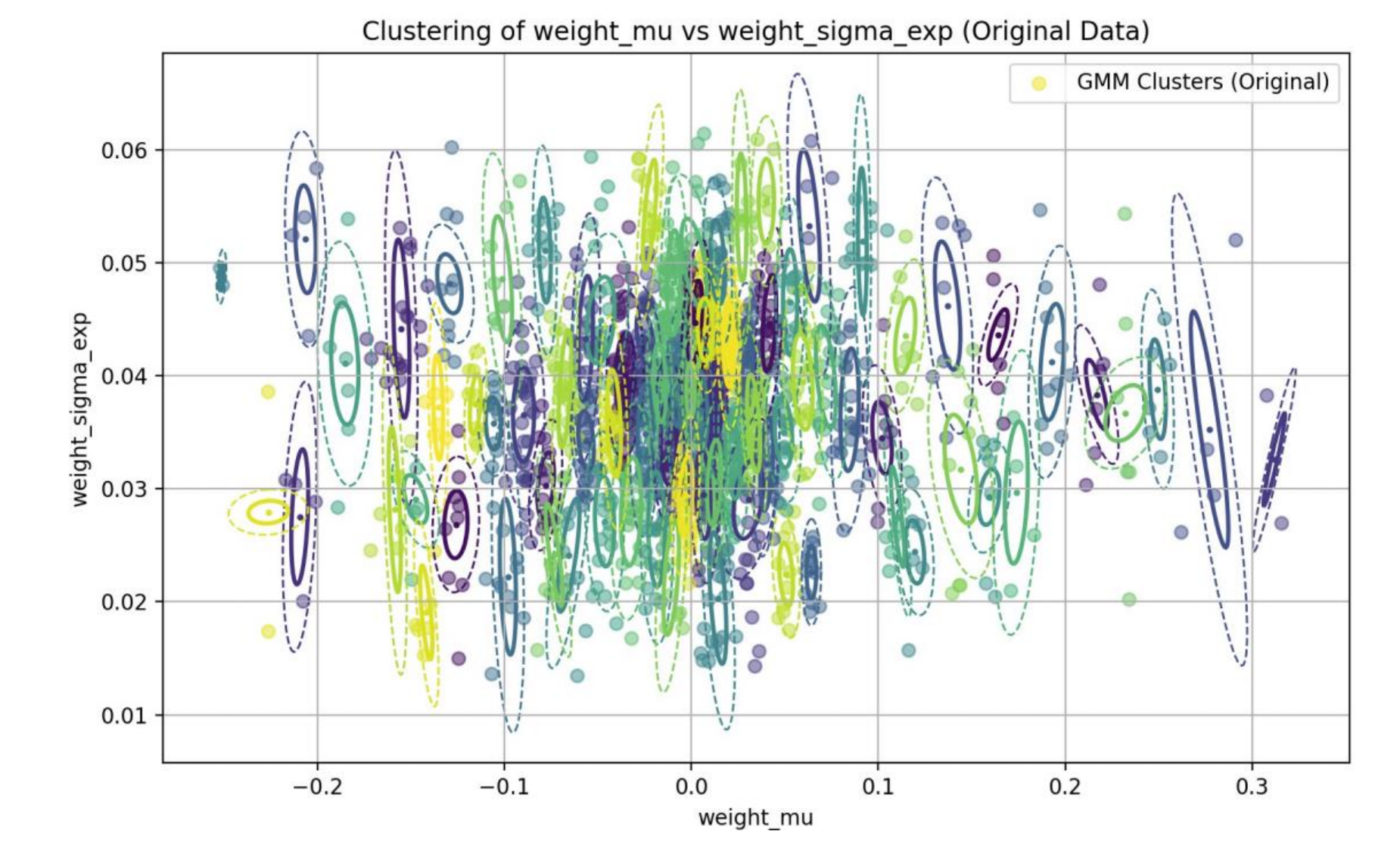}
\caption{Initial 2D Gaussian of the second convolutional layer of ResNet-18 in CIFAR-10 dataset. Different colors represent the different clusters. The solid line is the first variance area, and the dotted line is the second variance area.}
\label{fig:output}
\end{figure}

Additionally, the covariance resulting from the stochastic weight sharing is visualised in Figure~\ref{fig:output}. To clarify the behaviour of the model's uncertainty, we visualise the outputs of ResNet-18 on the CIFAR-10 dataset in Figure~\ref{fig:posteriorandoutput} (Right). This figure provides a view of the confidence distributions across all ten CIFAR-10 classes. %Significantly, the model demonstrates lower confidence in predicting the 'cat' class, evidenced by a median confidence of 0.79. It indicates a relatively higher level of uncertainty for the 'cat' class compared to the other classes.

\section{Stochastic Weight Initialisation}
Model initialisation always affects training and convergence. Here, we demonstrate the initial state of 2D Gaussian, including the visualisation of the covariance and center, as well as weights in different colours.
Figure~\ref{fig:output} visualises the 2D Gaussian of the second convolutional layer of ResNet-18 in CIFAR-10 dataset in the $\mu$-$\sigma$ space.

The ellipse weights, (the ones outside two standard deviations from any one cluster) are of crucial importance for inference, as most of them represent error points introduced by the Gaussian Mixture Model (GMM). To address this, we reassess their properties by reallocating them to multiple Gaussian distributions using alpha blending. 

%In the initial state, the shape, and orientation of the 2D Gaussian distribution in the Figure indicate that the variance along the $\sigma$ direction is obviously higher than that alone in the $\mu$ direction. The scale factor of $\mu$ is set to 0.1, while the scale factor of $\sigma$ is 0.01. This disparity in scaling units results from the stretching along the $\sigma$ direction.

\section{Training and Inference Time Comparison}
The overall training time will depend on the architecture and dataset employed, as well as the hardware used for training. Here, we give figures on the case of 2DGBNNs with the architecture of ResNet-18 trained for CIFAR-10, on a Tesla V100.  The inference was executed on an RTX 4070 (Laptop) with a batch size of 1024.

 \begin{table}[h!]
\centering
\caption{Training on NVIDIA Tesla V100 and Inference on NVIDIA RTX 4070 Laptop: A Comparative Analysis of Our 2DGBNNs Method and "Bayes by Backprop"}
\label{tab:training_inference_comparison}
\renewcommand{\arraystretch}{1.7}
\begin{tabular}{lcc}
\toprule
\textbf{Method}                    & \textbf{Training Time}                   & \textbf{Inference Time (CIFAR-10)} \\ 
\midrule
\textbf{Our Method (2DGBNNs)}      & Pre-training: 60--70 min                 & 1024 images: 12.75 sec            \\
                                   & Quantisation: 1.5--2 min                 & Total: 151.86 sec                 \\
                                   & Re-training: 30--35 min                  &                                    \\
                                   & \textit{Using existing pre-trained model: $\sim$half time} &                                    \\
\midrule
\textbf{Bayes by Backprop}         & Total: 5--5.5 hours                      & 1024 images: 73.62 sec            \\
(ResNet-18, CIFAR-10)              & (140--150 epochs, 10 samples, batch size 256) & Total: 882.41 sec                 \\
\bottomrule
\end{tabular}
\end{table}

We notice, that training is a one-off cost and that the main advantages will be in the computational time at inference time. From the Table.\ref{tab:training_inference_comparison}, we compared training and inference times between Our Method (2DGBNNs) and Bayes by Backprop for ResNet-18 on CIFAR-10. Our method significantly reduces training time (50–70 minutes pre-training, 1.5–2 minutes quantization, 30–35 minutes re-training) compared to 5–5.5 hours for Bayes by Backprop. Additionally, inference time is much faster with 2DGBNNs (151.86 seconds) versus Bayes by Backprop (882.41 seconds), demonstrating the efficiency of the multi-stage training approach.

\section{Out of Distribution Detection Task}

\begin{table}[h]
\centering
\renewcommand{\arraystretch}{1.7}
\caption{OOD Detection Results: ResNet-18-2DGBNN trained on CIFAR-10, evaluated on the SVHN dataset.}
\label{tab:ood_results}
\begin{tabular}{lc}
\hline
\textbf{Metric}                  & \textbf{Value}                \\ \hline
AUROC (Area Under ROC Curve)     & 0.8867                                         \\
AUPR (Area Under Precision-Recall Curve) & 0.8397                                          \\
FPR85 (False Positive Rate at 85\% True Positive Rate) & 0.2932                                      \\ \hline
\end{tabular}
\end{table}
In this experiment, we evaluated the performance of our model on the Out-Of-Distribution (OOD) detection task. The model, ResNet-18-2DGBNN, was trained on the CIFAR-10 dataset, with a compression ratio of 99\% and a parameter count of 0.018M. For OOD detection, we used the SVHN (Street View House Numbers) dataset. The result is shown in the Table \ref{tab:ood_results}. It achieved an AUROC (Area Under the ROC Curve) of 88.67\%. Additionally, the AUPR (Area Under the Precision-Recall Curve) reached 83.97\%, And FPR85 (False Positive Rate at 85\% True Positive Rate) got 29.32\%.
\section{Hyper-parameters Discussion}
\label{sec:hyperparameters}
The 2DGBNN involves several hyper-parameters. Among these, $\tau_w$, as shown in Algorithm \Rmnum{1}, line 6, is the most critical and sensitive parameter, playing a key role in the selection of outlier and inlier weights. To demonstrate the process of selecting $\tau_w$, we conducted an empirical experiment, the results of which are detailed in Table \ref{tab:tau_w_results}.

\begin{table}[h]
\centering
\renewcommand{\arraystretch}{1.7}
\caption{Experimental results for varying the value of \( \tau_w \) on 2DGBNNs with ResNet-20 architecture on CIFAR-10.}
\label{tab:tau_w_results}
\begin{tabular}{ccc ccccc}
\toprule
\textbf{\(\tau_w\)} & \textbf{Accuracy} & \textbf{NLL} & \textbf{ECE} & \textbf{Outliers} & \textbf{Ellipses} & \textbf{Gaussians} & \textbf{Parameters} \\
\midrule
0.1  & 91.14\% & 0.376  & 0.0197 & 223046 & 4501 & 2804 & 0.238M  \\
0.2  & 91.04\% & 0.303  & 0.037  & 14624  & 260  & 2387 & 0.020M  \\
0.4  & 90.53\% & 0.396  & 0.0279 & 101970 & 4369 & 120  & 0.111M  \\
0.6  & 89.78\% & 0.322  & 0.038  & 12211  & 3795 & 161  & 0.020M  \\
\bottomrule
\end{tabular}
\end{table}

We empirically selected the hyper-parameters for our methodology through a combination of grid search and trial-and-error over the training set. We performed an experimental evaluation of the 2DGBNNs with the ResNet-20 architecture on the CIFAR-10 dataset to study the effect of varying the value of $\tau_w = 0.2$.

These results demonstrate that as $\tau_w$ increases, fewer weights are classified as outliers, which reduces model complexity. However, this also slightly decreases accuracy and increases errors, with $\tau_w = 0.2$ achieving the best overall trade-off.

\section{Results with Deterministic Neural Networks}
In this section, we present the results of the experiments discussed in the main paper but obtained using deterministic neural networks, across three datasets and five architectures. These details are shown in the Table~\ref{tab:deterministic}. These deterministic networks are used as prior for the centring of GMMs in the main comparisons discussed in the main paper (Section 5.1). %stochastic neural networks, which also affects the 2DGBNNs training and inference. Combining these Tables with the main table, it drops acceptable performance for expanding the Bayesian framework to large-scale architecture and dataset.

\begin{table*}[!h]
\centering
\caption{Deterministic neural networks used as prior for the Initialisation 2DGBNN in the experiments discussed in Section 5.1.}
\renewcommand{\arraystretch}{1.4}
\resizebox{0.85\textwidth}{!}{
\begin{tabular}{@{}cccccc@{}}
\toprule
\textbf{Method} & Dataset  &
\textbf{Accuracy $\uparrow$} & 
\textbf{NLL} $\downarrow$ & 
\textbf{ECE} $\downarrow$ & {\textbf{\#Parameters (M: million)}}
\\
\hline
\multirow{4}{*}{ImageNet-1k} & \multicolumn{1}{|c}{ResNet-18} &   69.70& 1.265& 0.027&11.7M \\
& \multicolumn{1}{|c}{ResNet-50}& 76.10 &0.989 & 0.035 & 25.6M \\
& \multicolumn{1}{|c}{ResNet-101} &   77.30 & 0.936 & 0.936& 44.5M \\
& \multicolumn{1}{|c}{VIT-B-16} &   81.07& 0.856 &0.056& 86.0M \\
\hline
\multirow{4}{*}{CIFAR-100} & \multicolumn{1}{|c}{ResNet-18} &   77.10 &1.038 &0.114&11.7M \\
& \multicolumn{1}{|c}{ResNet-50}& 79.20&  0.950&  0.054 & 25.6M \\
& \multicolumn{1}{|c}{ResNet-101} &   80.02& 0.849& 0.095& 44.5M \\
& \multicolumn{1}{|c}{WRN-28-10} &   81.41& 0.766 &0.045&53.6M \\
\hline
\multirow{5}{*}{CIFAR-10} & \multicolumn{1}{|c}{ResNet-20} &   91.84&0.246& 0.031&0.27M \\
& \multicolumn{1}{|c}{ResNet-18} &   93.21& 0.201& 0.022&11.7M \\
& \multicolumn{1}{|c}{ResNet-50}& 94.81& 0.211& 0.010 &25.6M \\
& \multicolumn{1}{|c}{ResNet-101} &   94.70& 0.849& 0.095& 44.5M \\
& \multicolumn{1}{|c}{WRN-28-10} &   95.92& 0.131& 0.010& 53.6M \\
\bottomrule
\end{tabular}
}
\label{tab:deterministic}
\end{table*}
Table \ref{tab:deterministic} displays the performance metrics for deterministic models used as initialization priors for the stochastic counterparts in our 2DGBNNs framework. For each of the three datasets—ImageNet-1k, CIFAR-100, and CIFAR-10—we report the accuracy, Negative Log Likelihood (NLL), Expected Calibration Error (ECE), and the number of parameters. The models evaluated include variations of ResNet (ResNet-18, ResNet-50, ResNet-101) and other architectures such as VIT-B-16 and WRN-28-10.

The table is structured to highlight the performance across multiple architectures and datasets, facilitating a direct comparison. Higher accuracies and lower NLL and ECE values indicate better model performance. For instance, VIT-B-16 on ImageNet-1k achieves an accuracy of 81.07\% with the lowest ECE of 0.056 among its dataset counterparts. Similarly, WRN-28-10 shows superior performance on CIFAR-10 with the highest accuracy of 95.92\% and a remarkably low ECE of 0.010. The number of parameters, reported in millions, also provides insight into the model complexity, ranging from 0.27M for ResNet-20 on CIFAR-10 to 86.0M for VIT-B-16 on ImageNet-1k.

The 2DGBNN models on ImageNet1k showed a decrease in accuracy, with our ResNet-50 configuration dropping from the benchmark high of 77.5\% to 75.10\%. This reduction was coupled with a substantial decrease in the number of parameters—from models requiring up to 25.6 million parameters to just 0.101 million. Despite these changes, the increases in NLL and ECE were minimal and within acceptable ranges, indicating that the models maintain satisfactory predictive performance and calibration despite the reduced complexity. Similar trends were observed in the CIFAR-100 dataset, where our WRN-28-10 model's accuracy decreased from 81.41\% to 80.5\%, while significantly reducing the parameter count to only 0.045 million from 53.6 million. The NLL and ECE metrics, although slightly elevated, remained competitive, affirming the effective uncertainty estimation capabilities of the models despite their reduced complexity.

% \iffalse
% \section{Independent Gaussian Assumptions and Bayesian Neural Network Adaptation}

% \label{sec:ex_details}
% \subsection{Proof of the Algorithm in Stage 3 with Independent Gaussians}

% In \textbf{Stage 3} of the algorithm, we assume that the Gaussians (resulting from the Gaussian Mixture Model, GMM) are independent of each other. This is a weaker assumption compared to traditional Bayesian Neural Networks (BNNs), which assume that all individual weights are independent. By clustering weights into Gaussians, we allow for dependencies within clusters, thus making our assumption less restrictive.

% Below, we provide a mathematical justification of the algorithm, demonstrating that it is valid under the assumption of independent Gaussians.

% \subsection{Assumptions}
% \begin{enumerate}
%     \item \textbf{Independent Gaussians:} We assume that the Gaussians in the GMM are independent of each other.
%     \item \textbf{BNN Weight Independence:} Outliers weights are independent.
%     \item \textbf{Weaker Assumption:} By allowing dependencies within clusters (Gaussians), we make a weaker assumption than traditional BNNs.
% \end{enumerate}
% \fi
\section{Additional Background}

\paragraph{KL Divergence between Gaussians}

The KL divergence between two Gaussian distributions \( q(w) = \mathcal{N}(w | \mu_q, \sigma_q^2) \) and \( p(w) = \mathcal{N}(w | \mu_p, \sigma_p^2) \) is given by:
\begin{equation}
\text{KL}(q(w) \| p(w)) = \frac{1}{2} \left( \frac{\sigma_q^2}{\sigma_p^2} + \frac{(\mu_p - \mu_q)^2}{\sigma_p^2} - 1 + \ln \frac{\sigma_p^2}{\sigma_q^2} \right)
\end{equation}
This formula can be used to compute the KL divergence terms in the ELBO expression for both inliers and outliers.

\subsubsection*{Expected Log-Likelihood Computation}

The expected log-likelihood term involves an expectation over the variational posterior:
\begin{equation}
\mathbb{E}_{q(\mathbf{w})} \left[ \log p(\mathbf{y} | \mathbf{X}, \mathbf{w}) \right] = \int q(\mathbf{w}) \log p(\mathbf{y} | \mathbf{X}, \mathbf{w}) \, d\mathbf{w}
\end{equation}
In practice, this integral is intractable and is approximated using Monte Carlo sampling. 

\section{Wasserstein-based 2D Gaussian Merging}
In the context of comparing two Gaussian distributions, the Wasserstein distance provides a meaningful way to measure the distance between probability distributions. Specifically, for two 2D Gaussian distributions \(\mathcal{N}(\mu_1, \Sigma_1)\) and \(\mathcal{N}(\mu_2, \Sigma_2)\), where \(\mu_1, \mu_2\) are the means and \(\Sigma_1, \Sigma_2\) are the covariance matrices, the Wasserstein-2 distance is given by the following formula:

\begin{equation}
W_2(\mathcal{N}(\mu_1, \Sigma_1), \mathcal{N}(\mu_2, \Sigma_2))^2 = ||\mu_1 - \mu_2||^2_2 + \text{Tr}(\Sigma_1 + \Sigma_2 - 2(\Sigma_1^{1/2} \Sigma_2 \Sigma_1^{1/2})^{1/2})
\end{equation}

Where \(||\mu_1 - \mu_2||_2\) is the Euclidean distance between the means of the two distributions,
\text{Tr} is the trace operator, which sums the diagonal elements of a matrix,
\(\Sigma_1^{1/2}\) refers to the matrix square root of the covariance matrix \(\Sigma_1\).
\par
The Wasserstein distance threshold is set at \(1.5 \times 10^{-7}\), with further discussion on the \ref{merging}. If the distance between Gaussian components falls below this, gradient information is further considered, ensuring a precise analysis of component similarity.
\par
This method ensures that the newly formed Gaussian component accurately reflects the collective distribution characteristics of the initial components while maintaining minimal internal variation.
\par
\section{Sub-module Discussion and Hyper-parameter Configuration}

% \AP{embded this paragraph removed from the main text here if needed:  Data augmentation techniques such as random cropping and horizontal flipping were applied to the CIFAR and ImageNet1k datasets to prevent overfitting. }

\subsection{Effectiveness of 2D Gaussian Merging Discussion}
\label{merging}
In this section, we discuss the effectiveness of the merging of 2D Gaussian distributions in our method. Experimental results analysing its effect are listed in Table \ref{tab:results}. 
For the ResNet20 model on the CIFAR-10 dataset, after merging the 2D Gaussian, the accuracy improved from 88.77\% to 91.02\%, the NLL decreased from 0.3792 to 0.3066, and the ECE reduced from 0.0500 to 0.0374. For the ResNet18 model on the CIFAR-100 dataset, although the improvement is smaller, the accuracy still increased from 74.50\% to 74.63\%, and the NLL decreased from 1.0555 to 1.0535. These results demonstrate the effectiveness of our 2D Gaussian merging method in enhancing model performance and calibration.

\begin{table}[htbp]
\centering
\caption{Performance of ResNet20 on CIFAR-10 and ResNet18 on CIFAR-100 before and after merging Gaussian distributions.}
\begin{tabular}{lccccc}
\toprule
\textbf{Model} & \textbf{Dataset} & \textbf{Method} & \textbf{Accuracy (\%)} & \textbf{NLL} & \textbf{ECE} \\
\midrule
\multirow{2}{*}{ResNet20} & \multirow{2}{*}{CIFAR-10} & Before Merging Gaussian & 88.77 & 0.3792 & 0.0500 \\
 &  & After Merging Gaussian & 91.02 & 0.3066 & 0.0374 \\
\midrule
\multirow{2}{*}{ResNet18} & \multirow{2}{*}{CIFAR-100} & Before Merging Gaussian & 74.50 & 1.0555 & 0.0391 \\
 &  & After Merging Gaussian & 74.63 & 1.0535 & 0.0400 \\
\bottomrule
\end{tabular}
\label{tab:results}
\end{table}

\label{hyper}
\subsection{Data Augmentation}

To improve the robustness and generalization of our models, we applied a series of data augmentation techniques during training on the ImageNet-1K, CIFAR-100, and CIFAR-10 datasets. Table~\ref{tab:data-augmentation} summarises the specific augmentation methods used for each dataset.
For the ImageNet-1K dataset, we applied random resized cropping to obtain images of size $224 \times 224$ pixels. This was followed by random horizontal flipping with a probability of 50\% to augment the dataset with mirrored images. Color jittering was used to adjust the brightness, contrast, saturation, and hue of the images with factors of 0.4, 0.4, 0.4, and 0.1, respectively. The images were then converted to tensors and normalised using the standard mean and standard deviation values for ImageNet.
\begin{table}[htbp]
\centering
\caption{Summary of data augmentation techniques applied to each dataset.}
\label{tab:data-augmentation}
\begin{tabular}{ll}
\toprule
\textbf{Dataset} & \textbf{Data Augmentation Techniques} \\
\midrule
ImageNet-1K &
\begin{tabular}[t]{@{}l@{}}
Random resized crop to $224 \times 224$ pixels, \\
Random horizontal flip, \\
Color jittering (brightness=0.4, contrast=0.4, \\
saturation=0.4, hue=0.1), \\
Conversion to tensor, \\
Normalization
\end{tabular} \\
\addlinespace
CIFAR-100 &
\begin{tabular}[t]{@{}l@{}}
Conversion to tensor, \\
Padding of 4 pixels (reflection mode), \\
Random crop to $32 \times 32$ pixels, \\
Random horizontal flip, \\
Conversion to tensor, \\
Normalization
\end{tabular} \\
\addlinespace
CIFAR-10 &
\begin{tabular}[t]{@{}l@{}}
Random crop to $32 \times 32$ pixels, \\
Random horizontal flip, \\
Conversion to tensor, \\
Normalization
\end{tabular} \\
\bottomrule
\end{tabular}
\end{table}
In the case of the CIFAR-100 dataset, we started by converting the images to tensors. We then padded the images with 4 pixels on each side using reflection mode to preserve edge information. After padding, we performed a random crop to $32 \times 32$ pixels, followed by random horizontal flipping to introduce mirror variations. The images were converted back to tensors and normaliseded accordingly.

For the CIFAR-10 dataset, the augmentation process involved a random crop to $32 \times 32$ pixels, which helps in teaching the model to be invariant to translations. We also applied random horizontal flipping to include mirrored versions of the images. Finally, the images were converted to tensors and normalised to standardise the input data.

\subsection{Hyperparameters for training the deterministic network}
In our experiments, similar to those conducted by previous researchers, we standardised the hyperparameters across all models in the initial stage. This approach was applied to various models including ResNet-18, ResNet-50, ResNet-101, and VIT. The hyperparameters used are as follows:
\begin{table}[H]
    \centering
     \renewcommand{\arraystretch}{1.2} % Adjusts the row height
    \begin{tabular}{ccc}
        \toprule
        \textbf{Hyperparameter} & \textbf{Variable} & \textbf{Default Value} \\
        \midrule
        Batch Size & \texttt{-b} & 256 \\
        Warm-up Phases & \texttt{-warm} & 2 \\
        Learning Rate & \texttt{-lr} & 0.1 \\
        Resume Training & \texttt{-resume} & \texttt{False} \\
        Total Epochs & \texttt{-EPOCH} & 250 \\
        Milestones & \texttt{-MILESTONES} & [30, 60, 90, 120, 150, 200] \\
        Weight Decay & \texttt{-MultiStepLR} & 5e-4 \\
        Training Mean & \texttt{--TRAIN\_MEAN} & (0.5071, 0.4865, 0.4409) \\
        Training Std & \texttt{-TRAIN\_STD} & (0.2673, 0.2564, 0.2761) \\
        \bottomrule
    \end{tabular}
    \caption{Summary of hyperparameters in the neural network training configuration}
    \label{tab:hyperparameters}
\end{table}
The table summarizes the standardised hyperparameters used across all models in our neural network training configurations, mirroring settings from previous research. It outlines common parameters such as batch size, warm-up phases, learning rate, and total epochs, alongside specific settings like weight decay and learning rate milestones. % These parameters are chosen to balance computational efficiency and model performance, with considerations for preventing overfitting and ensuring stable learning progression. Additionally, data normalization parameters (mean and standard deviation) are specified to maintain input consistency and aid in effective model training.

\subsection{Discussion the Initialisation of $\sigma$}

We experimented with several different initialisation methods for our models, including initialisation via a specific function as detailed by Lee et al., random generation, and Gaussian distribution, among others.
According to our experimental results, we ultimately adopted the following initialisation methods for our neural network parameters.

For the weight parameter $\text{weight\_sigma}$, we used the Xavier uniform initialisation with a gain of 0.01, defined as:
\begin{equation}
\mathbf{W} \sim \mathcal{U}\left(-\frac{g}{\sqrt{n_{\text{in}} + n_{\text{out}}}}, \frac{g}{\sqrt{n_{\text{in}} + n_{\text{out}}}}\right)
\end{equation}
where $g = 0.01$, $n_{\text{in}}$ is the number of input units, $n_{\text{out}}$ is the number of output units, and $\mathcal{U}(a, b)$ denotes a uniform distribution between $a$ and $b$. 
\par
For the bias parameter $\text{bias\_sigma}$, we initialised it using a normal distribution with a mean of 0.0 and a standard deviation of 0.001:
\begin{equation}
   \mathbf{b} \sim \mathcal{N}(0, 0.001^2) 
\end{equation}
In our neural network, we assign different learning rates to different parameters. Table~\ref{tab:hyperparameters_lr} summarises these hyperparameters.

\begin{table}[h]
\centering
\caption{Summary of Hyperparameters Used in Our Experiments}
\label{tab:hyperparameters_lr}
\renewcommand{\arraystretch}{1.2}
\begin{tabular}{lcl}
\toprule
\textbf{Hyperparameter} & \textbf{Symbol} & \textbf{Value} \\
\midrule
Learning rate for weight and bias $\mu$ & $\eta_{\text{weight\_mu}}$ & $1 \times 10^{-4}$ \\
Learning rate for weight and bias $\sigma$ & $\eta_{\text{weight\_sigma}}$ & $1 \times 10^{-2}$ \\

\bottomrule
\end{tabular}
\end{table}

We set the learning rates for the parameters as follows: the learning rates for \textbf{weight\_mu} and \textbf{bias\_mu} are both $\eta_{\text{weight\_mu}} = 1 \times 10^{-4}$; the learning rates for \textbf{weight\_sigma} and \textbf{bias\_sigma} are both $\eta_{\text{weight\_sigma}} = 1 \times 10^{-3}$.

\subsection{Hyper-parameters for training Bayesian Neural Network}
In our experiments, we utilise several hyperparameters that are crucial for the performance and convergence of our Bayesian neural network (BNN) model. These hyperparameters are carefully selected based on empirical studies to balance computational efficiency and model accuracy. Table~\ref{tab:hyperparameters} summarises the hyperparameters used in our experiments.

\begin{table}[h]
\centering
\caption{Summary of Hyperparameters Used in Our Experiments}
\label{tab:hyperparameters}
\renewcommand{\arraystretch}{1.2}
\begin{tabular}{lcl}
\toprule
\textbf{Hyperparameter} & \textbf{Symbol} & \textbf{Value} \\
\midrule
Initial learning rate for $\mu$ & $\eta_{\mu}$ & $1 \times 10^{-4}$ \\
Number of epochs & $E$ & 200 \\
Number of clusters in KMeans & $K$ & 6000 \\
Outlier threshold (weight value) & $T_w$ & $\pm 0.2$ \\
Outlier threshold (gradient percentile) & $P_g$ & Top 1\% \\
Minimum samples per cluster & $N_{\text{min}}$ & 20 \\
Learning rate in BNN training & $\eta_{\text{BNN}}$ & $1 \times 10^{-5}$ \\
Number of samples in predictive function & $N_s$ & 30 \\
Number of bins in ECE computation & $N_b$ & 15 \\
Wasserstein distance threshold & $T_W$ & $1 \times 10^{-2}$ \\
Mahalanobis distance threshold & $T_M$ & 5.991 \\
Number of nearest Gaussians & $k$ & 5 \\
\bottomrule
\end{tabular}
\end{table}

In the KMeans clustering algorithm, we initially use $K = 2000$ clusters. %This relatively large number of clusters allows us to capture the intricate structures in the weight distributions of the neural network layers. By having more clusters, we can model the weight distributions more precisely, which is beneficial for the subsequent Bayesian inference steps.
Outliers in the weight values are identified using a threshold $T_w = \pm 0.2$. Any weight value exceeding this threshold is considered an outlier. This threshold is chosen based on the empirical distribution of the weights after initial training. Similarly, outliers in the gradients are identified by selecting the top $P_g = 1\%$ of gradient magnitudes. %These thresholds help in isolating anomalous weights and gradients that could adversely affect model performance.

A minimum cluster size of $N_{\text{min}} = 30$ is enforced to ensure statistical significance in the clustering results. Clusters with fewer than $N_{\text{min}}$ samples are considered invalid and their associated weights are treated as outliers. This prevents the model from being influenced by clusters that may represent noise or insignificant patterns.

During the BNN training, we use a learning rate of $\eta_{\text{BNN}} = 1 \times 10^{-5}$, which is lower than the initial learning rate used in the preliminary training. The smaller learning rate is necessary to accommodate the Bayesian updates and to ensure that the posterior distributions over the weights converge properly.

In the predictive function, we draw $N_s = 30$ samples from the posterior distribution to estimate the predictive mean and uncertainty. %This number of samples provides a good trade-off between computational cost and the accuracy of the uncertainty estimates.
The Expected Calibration Error (ECE) is computed using $N_b = 15$ bins. %This number of bins is sufficient to capture the calibration of the model across different confidence levels without introducing excessive variance due to small sample sizes in each bin.
The Mahalanobis distance threshold $T_M = 5.991$ corresponds to the chi-squared distribution value with 2 degrees of freedom at the 95\% confidence level. The Mahalanobis distance for a data point $x$ with respect to a Gaussian distribution with mean $\mu$ and covariance $\Sigma$ is calculated as:
\begin{equation}
D_M^2 = (x - \mu)^\top \Sigma^{-1} (x - \mu).
\end{equation}
Points with a Mahalanobis distance greater than $T_M$ are considered outliers. %This statistical measure helps in robustly identifying data points that do not conform to the expected distribution.

In handling outliers, we consider the $k = 5$ nearest Gaussian components for each outlier point. This allows us to reassign outlier weights to the most probable Gaussian components based on their proximity in the parameter space.

%The careful selection of these hyperparameters is based on extensive empirical evaluation. Adjusting these values can have significant effects on model performance. For instance, increasing the number of clusters $K$ can lead to a more accurate representation of the weight distribution but at the cost of increased computational complexity. 

% \bibliographystyle{apalike} % 选择引用格式
% \bibliography{iclr2025_conference}
\vfill
\end{document}